\documentclass[conference]{IEEEtran}
\IEEEoverridecommandlockouts
\usepackage{cite}

\usepackage{soul}
\usepackage{pdfpages}
\usepackage{color}
\usepackage{xcolor}
\usepackage{url}
\usepackage{listings}   
\usepackage{amsfonts}
\usepackage{nicefrac}
\usepackage{microtype}
\usepackage{amsmath}
\usepackage{graphicx}
\usepackage{subfigure}
\usepackage{algorithm}
\usepackage{algpseudocode}
\usepackage{multirow}
\usepackage{mathtools}
\usepackage{balance}
\usepackage{bm}
\usepackage{dsfont}
\usepackage{accents}
\usepackage{booktabs}

\usepackage{color, colortbl}

\begin{document}

\title{Reliable and Fast Recurrent  Neural  Network  Architecture  Optimization
}

\author{\IEEEauthorblockN{Andr\'es Camero}
\IEEEauthorblockA{\textit{German Aerospace Center (DLR)} \\
Munich, Germany \\
andres.camerounzueta@dlr.de}
\and
\IEEEauthorblockN{Jamal Toutouh}
\IEEEauthorblockA{\textit{ITIS Software, Universidad de M\'alaga} \\
Malaga, Spain \\
jamal@lcc.uma.es}
\and
\IEEEauthorblockN{Enrique Alba}
\IEEEauthorblockA{\textit{ITIS Software, Universidad de M\'alaga} \\
Malaga, Spain \\
eat@lcc.uma.es}
}

\maketitle

\begin{abstract}
This article introduces Random Error Sampling-based Neuroevolution (RESN), a novel automatic method to optimize recurrent neural network architectures. RESN combines an evolutionary algorithm with a training-free evaluation approach. 
The results show that RESN achieves state-of-the-art error performance while reducing by half the computational time.
\end{abstract}


\vspace{-0.1cm}
\section{Introduction}

A recurrent neural network (RNN) is a class of artificial neural network that has feedback connections between nodes. Thanks to this recurrence, RNNs are particularly good for tackling time-dependent (or sequential) problems. But, due to this same feature (i.e., the recurrence), they are hard to train~\cite{Pascanu2013}: small changes on its components may produce a big performance deviation~\cite{Pascanu2013}.

Several alternatives have been proposed to deal with the sensitivity of RNN design, including specific node architectures (e.g., GRU~\cite{cho2014gru} and LSTM~\cite{Hochreiter1997}), for their fully automatic design~\cite{Liang:2018:EAS:3205455.3205489}. However, in spite of great improvements made so far~\cite{Ojha2017}, finding the \textit{best} design is still an open issue. Particularly, optimizing an RNN is (time and computation wise) a very demanding task. Therefore, a \emph{low-cost} approach is desirable.

This article summarizes our previous work~\cite{camero2019optlowcost}, in which we proposed a fast and accurate method to optimize the architecture of an RNN. Particularly, we have introduced Random Error Sampling-based Neuroevolution (RESN): a $(\mu + \lambda)$ evolutionary algorithm (EA) that uses MRS~\cite{camero2018lowcost} to guide its search, and Adam~\cite{kingma2014adam} to train the final solution. We have empirically validated our proposal on four prediction problems, and compared our technique to training-based architecture optimization techniques, neuroevolutionary approaches, and human expert designed solutions. The results of our experiments show that RESN achieves state-of-the-art performance, while reducing by half the overall time.

\section{Random Error Sampling-based Neuroevolution}

Given an input \emph{X}, an output \emph{Y}, a search space of RNN architectures (ARQ) and look back (or time steps, LB), we considered the problem of maximizing the \emph{fitness(X, Y)}=$p_t$, the probability of finding a set of weights whose performance is below the defined threshold (i.e., MRS~\cite{camero2018lowcost}).

To solve the stated problem, we proposed RESN, a ($\mu+\lambda$) EA. At a glance, the \emph{population} is a set of solutions, where each \emph{solution} represents an RNN architecture. The initial population is randomly set by the \textbf{Initialize} function. Later, the population is assessed by the \textbf{Evaluate} function, that computes $p_t$ for each solution.
Then, the population is evolved by the selection, mutation, evaluation, replacement, and self-adjustment operations. Once the termination criterion is met, i.e., the number of \emph{evaluations} is greater than the budget (\emph{max\_evaluations}), the \textbf{Best} solution (i.e., the one with the highest $p_t$) is selected and trained using Adam~\cite{kingma2014adam} for a predefined number of \emph{epochs}.

\setlength{\intextsep}{1pt}

\begin{figure}[!ht]
    \centering
    \includegraphics[width=0.6\columnwidth]{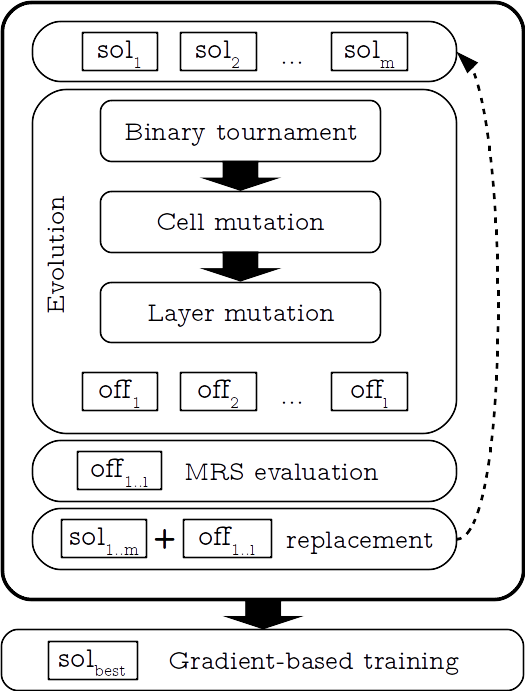}
    \caption{The global scheme of RESN}
    \label{fig:resn}
\end{figure}

Fig.~\ref{fig:resn} summarizes RESN. The proposed approach combines evolutionary computation and machine learning techniques to optimize the architecture and the weights of an RNN.

\section{Experimental Study}

We tested RESN on four problems: the sine wave, the waste generation prediction problem~\cite{Ferrer2018Bio}, the coal-fired power plant flame intensity prediction problem~\cite{ororbia2019investigating}, and the EUNITE load forecast problem~\cite{chen2004load}. Also, we proposed four experiments: 
\textbf{E1}: RESN vs. Gradient-based Architecture Optimization, we compared RESN against (i) a modified version of the same algorithm (i.e., a version that replaced the MRS by the results of Adam training), (ii) the \emph{Short training}~\cite{camero2018wastepred} algorithm, and (iii) a \emph{Random Search} algorithm.
\textbf{E2}: RESN vs. Neuroevolution, we benchmarked RESN against EXALT~\cite{elsaid2019evolving}.
\textbf{E3}: Optimization Time, we logged the execution times.
And \textbf{E4}: RESN vs. Expert Design, we compared our results against the winner~\cite{chen2004load} of the EUNITE load forecast competition and to recent solutions to the same challenge~\cite{lang2018short}.

Summarizing \textbf{E1}, (E1.i) the results of RESN exceed GDET (i.e., a modified version of RESN that uses training results to evaluate the performance of a solution). On average, RESN obtained a MAE of 0.105 while GDET got a 0.142 (Wilcoxon rank-sum test $p$-value equals to 0.001: significant). Then, (E1.ii) RESN was compared to \emph{Short Training}~\cite{camero2018wastepred}. The results show that there is no significant improvement in the error (i.e., Wilcoxon rank-sum test $p$-value equal to 0.665), however, RESN cut by half the optimization time (i.e., test \textbf{E3}). Table~\ref{table:results-rq1-ii} presents the results, where \emph{MAE} stands for the MAE of the final solution, and \emph{Time} for the total time (i.e., optimization and training of the final solution) in minutes.

\begin{table}[!ht]
\renewcommand{\arraystretch}{0.92}
    \caption{E1.ii. RNN optimization in the Waste generation prediction problem, a comparison between Short training and RESN}
    \vspace{-0.2cm}
	\centering
    \begin{tabular}{ lrrrr }
    \toprule
    & \multicolumn{2}{c}{Short Training} & \multicolumn{2}{c}{RESN} \\
    \cline{2-3}  \cline{4-5}
    & MAE & Time [min] & MAE & Time [min] \\
    \hline
        Mean    & 0.073 & 97 & 0.079 & 51 \\
         Median  & 0.073  & 70  & 0.073 & 45 \\
         Max     & 0.076  & 405   & 0.138 & 103 \\
         Min     & 0.071  & 33  & 0.069 & 40 \\
         SD      & 0.001  & 75 & 0.017 & 13 \\

    \bottomrule
	\end{tabular} \\    
    \label{table:results-rq1-ii}
\end{table}

To conclude E1, we compared RESN against random search (E1.iii). Despite the relatively good error performance of random search, with an average of 0.091, the Wilcoxon rank-sum test revealed that RESN (and \emph{Short training}) beats random search ($p$-values are 0.017 and 0.002 respectively). Moreover, random search took nearly 50x the time of RESN (!). Thus, RESN is a fast and accurate approach (test \textbf{E3}). 

Later, (E2) we compared RESN against neuroevolution. Summarizing, RESN improved EXALT by ten times (Wilcoxon rank-sum test $p$-value is 2.958e-06). Table~\ref{table:results-rq2} depicts the mean square error (MSE) of the solution obtained by RESN and EXALT~\cite{elsaid2019evolving} in the coal-fire power plan problem. 

\begin{table}[!ht]
\renewcommand{\arraystretch}{0.9}
    \caption{E2. RESN vs. EXALT in the coal-fire power plant problem (MSE)}
	\centering
	\vspace{-0.2cm}
    \begin{tabular}{ lrr }
    \toprule
    Fold & EXALT & RESN \\
    \hline
    0 & 0.028749 & 0.001541 \\
    1 & 0.031769 & 0.006536 \\
    2 & 0.023095 & 0.003821 \\
    3 & 0.019229 & 0.000570 \\
    4 & 0.023170 & 0.003336 \\
    5 & 0.036091 & 0.000617 \\
    6 & 0.012879 & 0.017061 \\
    7 & 0.019358 & 0.004032 \\
    8 & 0.018151 & 0.001912 \\
    9 & 0.019475 & 0.013996 \\
    10 & 0.030016 & 0.006120 \\
    11 & 0.031207 & 0.002942 \\
    \hline
    Average & 0.024432 & 0.005208 \\
    \bottomrule
	\end{tabular} \\    
    \label{table:results-rq2}
\end{table}

Finally (test \textbf{E4}), we compared RESN against human expert designed solutions in the EUNITE load forecast problem. Table~\ref{table:results-rq4} presents the results. The column SVM corresponds to~\cite{chen2004load}, the winner of the ``Electricity Load Forecast using Intelligent Adaptive Technologies'' competition organized by EUNITE, and the other columns, i.e., BP, RBF, SVR, NNRW, KNNRW, and WKNNRW, correspond to the results presented in~\cite{lang2018short}. NA stands for \emph{Not Available}. From the results, we concluded that the performance of RESN is comparable to a human expert, i.e., ``the error of the best solution found is as good as the best solution proposed by the experts''.

\begin{table}[!ht]
\setlength{\tabcolsep}{3pt}
\renewcommand{\arraystretch}{0.92}
    \caption{E4 results (MAPE). RESN . Expert design}
	\centering
	\vspace{-0.2cm}
    \begin{tabular}{ lrrrrrrrr }
    \toprule
            & 	SVM     & 	BP  & 	RBF & 	SVR & NNRW  & KNNRW & WKNNRW & 	RESN \\
    \hline
    Mean    & 	2.88   & NA    & NA    & NA    & NA    & NA    & NA    & 	2.28 \\ 
    Median  & 	2.95   & NA    & NA    & NA    & NA    & NA    & NA    & 	2.24 \\ 
    Max     & 	3.48   & NA    & NA    & NA    & NA    & NA    & NA    & 	3.27 \\ 
    Min     & 	1.95   & 1.45 & 1.48 & 1.45 & 1.44 & 1.35 & 1.32 & 	1.37 \\ 
    SD     & 	0.01   & NA    & NA    & NA    & NA    & NA    & NA    & 	0.41 \\ 
    \bottomrule
	\end{tabular} \\    
    \label{table:results-rq4}
\end{table}

\section{Conclusions and Future Work}

As a summary of this study, we conclude that RESN (which is a training-free algorithm) is a competitive approach to RNN design. Particularly, it achieves a comparable error performance of training-based RNN techniques and neuroevolutionary approaches but considerably reduces the computational time, and it is as good as human expert designed solutions.

\bibliographystyle{IEEEtran}
\bibliography{bibliography}

\end{document}